\pdfoutput=1
\documentclass[sigconf]{acmart}

\AtBeginDocument{%
  \providecommand\BibTeX{{%
    \normalfont B\kern-0.5em{\scshape i\kern-0.25em b}\kern-0.8em\TeX}}}

\usepackage{algorithm}
\usepackage{algorithmic}

\usepackage{subcaption}
\usepackage[utf8x]{inputenc}
\usepackage{multirow}
\usepackage{amsmath} 
\usepackage{enumerate}

\setcopyright{none}
\renewcommand\footnotetextcopyrightpermission[1]{} 
\begin{document}

\title{PRE-NAS: Predictor-assisted Evolutionary Neural Architecture Search}


\author{Yameng Peng}
\affiliation{%
  \institution{RMIT University}
  \city{Melbourne}
  \state{Victoria}
  \country{Australia}
  \postcode{3000}
}
\email{yameng.peng@student.rmit.edu.au}

\author{Andy Song}
\affiliation{%
  \institution{RMIT University}
  \city{Melbourne}
  \state{Victoria}
  \country{Australia}
  \postcode{3000}
}
\email{andy.song@rmit.edu.au}

\author{Vic Ciesielski}
\affiliation{%
  \institution{RMIT University}
  \city{Melbourne}
  \state{Victoria}
  \country{Australia}
  \postcode{3000}
}
\email{vic.ciesielski@rmit.edu.au}

\author{Haytham M. Fayek}
\affiliation{%
  \institution{RMIT University}
  \city{Melbourne}
  \state{Victoria}
  \country{Australia}
  \postcode{3000}
}
\email{haytham.fayek@ieee.org}

\author{Xiaojun Chang}
\affiliation{%
  \institution{University of Technology Sydney}
  \city{Sydney}
  \state{New South Wales}
  \country{Australia}
  \postcode{2007}
}
\email{xiaojun.chang@uts.edu.au}

\renewcommand{\shortauthors}{Peng, et al.}

\begin{abstract}
Neural architecture search (NAS) aims to automate architecture engineering in neural networks. This often requires a high computational overhead to evaluate a number of candidate networks from the set of all possible networks in the search space during the search.  Prediction of the networks' performance can alleviate this high computational overhead by mitigating the need for evaluating every candidate network.  Developing such a predictor typically requires a large number of evaluated architectures which may be difficult to obtain.  We address this challenge by proposing a novel evolutionary-based NAS strategy, Predictor-assisted E-NAS (PRE-NAS), which can perform well even with an extremely small number of evaluated architectures.  PRE-NAS leverages new evolutionary search strategies and integrates high-fidelity weight inheritance over generations.  Unlike one-shot strategies, which may suffer from bias in the evaluation due to weight sharing, offspring candidates in PRE-NAS are topologically homogeneous, which circumvents bias and leads to more accurate predictions.  Extensive experiments on NAS-Bench-201 and DARTS search spaces show that PRE-NAS can outperform state-of-the-art NAS methods.  With only a single GPU searching for 0.6 days, competitive architecture can be found by PRE-NAS which achieves 2.40\% and 24\% test error rates on CIFAR-10 and ImageNet respectively.
\end{abstract}




\maketitle

\section{Introduction}

Despite the overwhelming success of deep learning, the high computational cost associated with model development remains a challenge in the field, especially in real-world applications which often require carefully constructing complicated structures such as ResNet \cite{Ref:04}, DenseNet \cite{Ref:47} and manually tuning the hyperparameters \cite{Ref:18,Ref:19,Ref:04,Ref:28,Ref:46}.  One approach to address this challenge is NAS (neural architecture search).  The aim is to automatically create a competitive neural network for a given task.  A fully-machine-designed neural network \cite{Ref:01} can achieve a test accuracy of 96.35\% on the CIFAR-10 dataset, in comparison with 96.54\% from DenseNet \cite{Ref:47}, a hand-designed neural network.  However, in this task, NAS requires 800 Tesla K40 GPUs parallel running for nearly one month.  This high computational cost compromises the saving on manual architecture design.  Hence, reducing of the computational costs is one of the key targets in NAS.

One way to reduce the cost in NAS is to cut down evaluation, for example, using fewer epochs or a small portion of data to train the networks \cite{Ref:48,Ref:21,Ref:08}.  However, this approach may lead to inadequate training, and therefore inaccurate results. Another approach, learning curve extrapolation methods \cite{Ref:12,Ref:26}, predicts the tendency of network optimisation based on early epochs in training.  Similarly, a surrogate model can be built to predict the performances of candidate networks \cite{Ref:40}.  Predictor-based evaluation methods require their own training, hence there is a need to sample networks from the search space with associated high computational cost.  Weights sharing provides a good alternative, in particular, the one-shot model \cite{Ref:15,Ref:17,Ref:10}, which treats all candidate networks as sub-networks of an over-parameterised super-network.  Sub-networks within the same search space can share weights, hence computational costs can be reduced.  However sharing weights between heterogeneous architectures is problematic, and can easily lead to incorrect ranking of the candidate networks \cite{Ref:49,Ref:50}.


Hence, this work addresses the balance between reducing computational cost and improving network ranking by proposing a predictor-assisted evolutionary architecture search algorithm (PRE-NAS).  It is driven by an evolutionary algorithm that is based on a population of candidate networks rather than one network. We introduce several new strategies to train predictors more effectively, especially with extremely limited training samples.  Firstly, elitist evolution is introduced to maintain a good pool of candidates.  Secondly, multi-mutation and a representative selection strategy are introduced.  These strategies can improve the training set for the predictor by heuristically increasing the number of mutations and sampling representative candidates from each generation. So the predictor can evaluate multiple candidate networks with no significant cost increase.  In addition, a high-fidelity weight inheritance is incorporated to reduce the computational cost further. 


The contributions of this work are summarised as follows:
\begin{enumerate}

    
    
    
    
    \item We propose a predictor-assisted evolutionary search algorithm (PRE-NAS) which outperforms several mainstream NAS algorithms on benchmark and real-world search spaces in terms of efficiency and performance.
    
    \item We introduce a representative selection strategy that can train a good performance predictor using an extremely limited number of training samples in the NAS context. Moreover, we apply a multi-mutation strategy to utilise the performance predictor maximally in the evolutionary NAS.
    
    
    \item We report a 2.40\% test error rate on CIFAR-10 and a 24\% top-1 test error rate on ImageNet (mobile setting), with only 0.6 GPU days to search.
    
    
\end{enumerate}

\section{Related Work}
\label{section2}
NAS is generally studied from three perspectives: search algorithm, search space, and network evaluation strategy.  Typical search algorithms include reinforcement learning (RL), evolutionary algorithm (EA), gradient-based and Bayesian optimisation (BO). Search space is continuous in gradient-based methods like DARTS \cite{Ref:10}. Thus, the search space and evaluation strategy are tightly coupled. In contrast, search space in evolutionary-based \cite{Ref:36,Ref:51,Ref:08,Ref:40,Ref:37,Ref:07} and reinforcement learning-based \cite{Ref:01,Ref:39} methods is usually discrete, offering better flexibility and compatibility for network evaluation strategies.  RL is more costly in comparison.  Hence we leverage the advantages of EA as the search algorithm in this NAS study.

There are many successful evolutionary or predictor-assisted NAS work \cite{Ref:07,Ref:08,Ref:27,Ref:36,Ref:37,Ref:40,Ref:41,Ref:51}. AmoebaNet \cite{Ref:08} adopts a regularised evolutionary algorithm called Aging Evolution.  It uses one mutation operation to generate a new architecture and discards the oldest architecture from the population in each search cycle.  However, it is not efficient.  Duplicate offspring are not checked so it is possible to see identical architectures in two different search cycles.  The oldest individual could be the best-performing one in the population, so adding more unnecessary cost onto the search.  PNAS \cite{Ref:40} adopts a sequential model-based optimisation (SMBO) search algorithm and a multi-layer perceptron (MLP) ensemble predictor.  PNAS search starts from shallow cell networks and progresses to complex ones.  During the architecture search, a predictor is trained to predict the performance of candidate networks without needing to train all of the candidate networks.  After evaluation, top-$k$ candidates are selected for further training \cite{Ref:40}.  This strategy needs to define a suitable $k$ which may change for different tasks.  

Besides, in order to calibrate the performance predictor during architecture search, we adopt weight inheritance training for a few representative candidate networks.  Unlike other work \cite{Ref:55,Ref:36} inheriting weights between variable-size of generation, our work adopts fixed-size weight inheritance.  The topology of parent and offspring networks are similar, hence helpful for function preservation during weight inheritance.
In PRE-NAS, we propose a percentile representative selection strategy that is more generalizable than the top-$k$ strategy.  PRE-NAS incorporates the flexibility of the EA, efficiency of performance prediction and weight inheritance to address the aforementioned issues in NAS. 

 \begin{figure}[!t]
    \begin{center}
  	\includegraphics[width=0.9\linewidth,height=0.28\textheight]{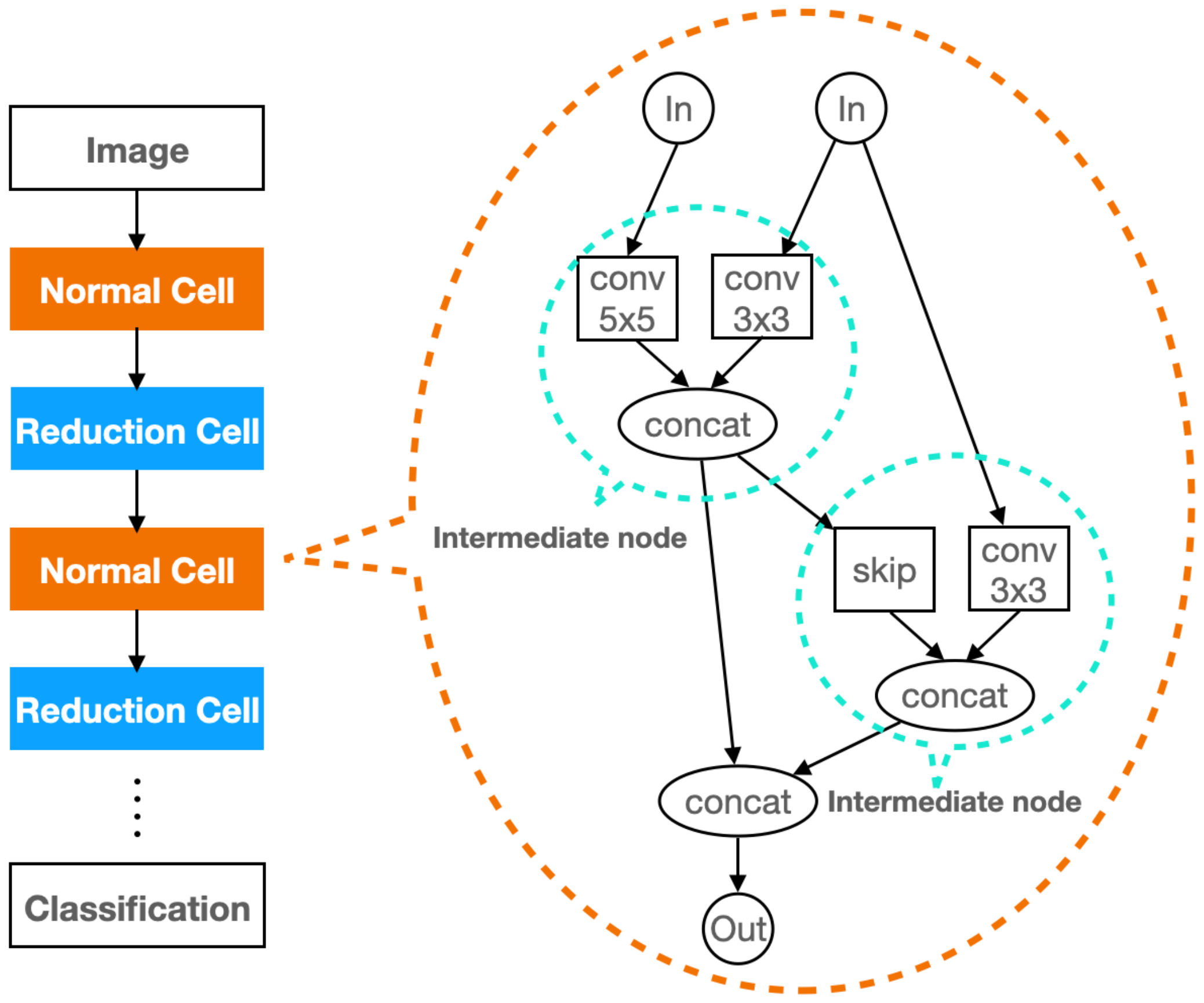}
  	\caption{Illustration of a cell-based network. Left-hand side demonstrates the complete stacking-constructed network. The micro-structure in the orange dash line represents a cell network. An inside circle represents an intermediate node.}
  	\label{cell}
  	\end{center}
\end{figure} 

\section{Predictor-assisted Evolutionary NAS}
\label{section3}
The details of PRE-NAS are described here, including the search space, search algorithm, multi-mutation, representative selection, performance predictor, and weight inheritance training.

\subsection{Search Space of Cell-based Networks}
\label{search_space}
The search space is defined by the specific task and the chosen representation.  Two search spaces, NAS-Bench-201 \cite{Ref:45} and DARTS \cite{Ref:10} are studied here.  They are both cell-based search spaces, as many well-performing networks have been manually designed based on this representation.  Fig. \ref{cell} is an illustration of the construction of a cell-based network by stacking repeated modules together~\cite{Ref:04,Ref:47}.  The shallow network in the oval callout on the right is called a Cell Network. Each cell network consists of several pre-defined operations. A complete network can be constructed by stacking these cells (Fig. \ref{cell} left).  The number in a stack, e.g. the depth of the network, depends on the difficulty of the target task.  Thus, the search algorithm only needs to focus on the microstructure of a cell network.

 \begin{figure*}[!t]
    \begin{center}
  	\includegraphics[width=0.9\linewidth,height=0.26\textheight]{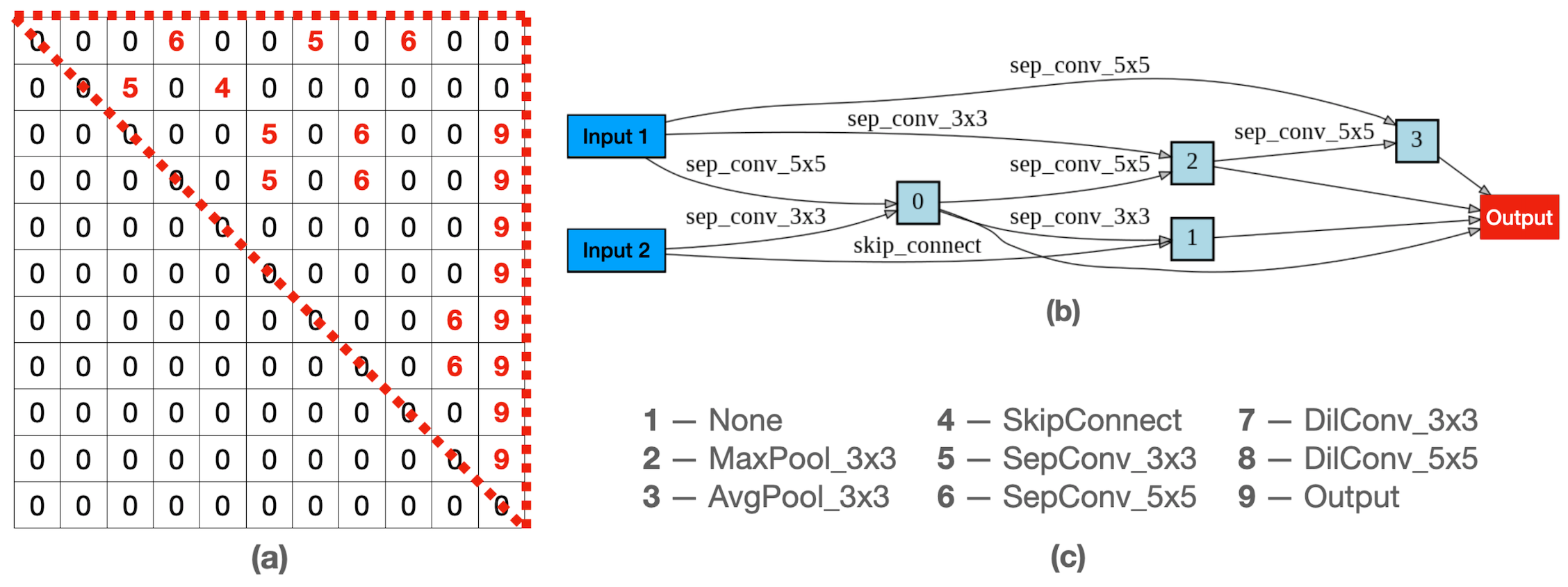}
  \caption{Illustration of upper triangular adjacency matrix representation for architecture encoding. The corresponding architecture is visualised on the top right demonstrating a cell network (b).  The list (c) on the bottom right shows the available network operations and their indices, which are the cell values of the matrix (a) on the left.}
  	\label{encoding}
  	\end{center}
\end{figure*} 

The search space of NAS-Bench-201 \cite{Ref:45} is relatively simple as it aims to provide a fair environment for comparison between NAS algorithms. The elements are four nodes (one input node, two intermediate nodes and one output node) and five pre-defined network operations (none, skip-connection, 1x1 convolution, 3x3 convolution, 3x3 average pooling) connecting these nodes. So the search space consists of $5^6 = 15625$ architectures.  Thus, it is possible to exhaustively train all $15625$ architectures for 200 epochs on three different image datasets and record their individual performance. Then the accuracy of a candidate network can be obtained by querying the records instead of training from scratch.

DARTS space appears much more recently  \cite{Ref:52,Ref:33,Ref:31,Ref:53,Ref:07}. It has a much larger and more complex search space than NAS-Bench-201. There are eight different operations available in the search space of DARTS: $3\times3$ and $5\times5$ separable convolutions, $3\times3$ and $5\times5$ dilated convolutions, $3\times3$ max pooling, $3\times3$ average pooling, skip connection and none.  Each cell network has 2 input nodes, 4 intermediate nodes and 1 output node, leading to $\prod_{i=1}^{4} \frac{(i+1)i}{2} \times 7^2 \approx 10^9 $ possible architectures. Thus, more complex and powerful neural architectures can be found in this search space.  This study adopts a similar setup as DARTS, except that DARTS searches for a Reduction Cell with a different structure from the Normal Cell.  Our Reduction Cell and Normal Cell have the same structure, similar to the strategy in PNAS \cite{Ref:40} and BONAS \cite{Ref:31}.

The encoding of the above two search spaces both use upper triangular adjacency matrices, as demonstrated in Fig. \ref{encoding}.  A cell network is represented as an upper triangular matrix where the number in a cell indicates the network operation connecting the two nodes represented in the row and in the column, similar to the encoding schemes in \cite{Ref:45,Ref:31}.  This matrix is also the input to our performance predictor.  Note that the output (No. 9) in the operation list is to demonstrate the encoding scheme,  but it is not involved in the actual architecture search, e.g, mutations or predictor training.

\begin{algorithm}[!t]
\caption{PRE-NAS}
\label{algo}
\begin{algorithmic}[1]
\REQUIRE Population size P, Search cycle C, Sample size S
\STATE $\mathit{population} \gets \emptyset$
\STATE $\mathit{history} \gets \emptyset$
\STATE $\mathit{Children} \gets \emptyset$
\WHILE {$population$  $< P$}
\STATE $\mathit{model.arch} \gets RandomGenerateArchitectures()$ 
\STATE $\mathit{model.valid\_accuracy} \gets Train(model.arch)$ 
\STATE Add $model$ to the $population$
\STATE Add $model$ to $history$
\ENDWHILE

\WHILE {$C$ not fulfilled}
\STATE $\mathit{Predictor}\gets$ training the predictor
\STATE $\mathit{candidates}\gets$ randomly sample $S$ architectures from the $population$

\STATE $\mathit{parent} \gets$ best-performing one in the $candidates$  

\WHILE {$mutation\_times$ not fulfilled}
\STATE $\mathit{child.arch} \gets$ $Mutate(parent.arch)$
\STATE $\mathit{child.valid\_accuracy} \gets$ $Predictor(child.arch)$
\STATE Add $child$ to $Children$
\ENDWHILE

\STATE $\mathit{Representatives} \gets$ select \textbf{few} representatives from $Children$

\FORALL{$child\in Representatives $}
\STATE $\mathit{child.valid\_accuracy} \gets Inheritance\_Train(child.arch)$
\STATE Add $child$ to $history$
\ENDFOR
\STATE Calculate \textbf{Spearman coefficient} between predicted and trained accuracy of  $\mathit{Representatives}$;
\STATE Enlarge $mutation\_times$ if \textbf{Spearman coefficient} higher than previous
\STATE Add the best-performing $child$ to the $population$
\STATE Remove the worst $architecture$ from the $population$
\ENDWHILE
\STATE \textbf{end}
\end{algorithmic}
\end{algorithm}

\subsection{Evolutionary Search Algorithm}

As the evolutionary search algorithms \cite{Ref:07,Ref:08,Ref:27, Ref:36,Ref:37,Ref:60,Ref:62} have shown competitiveness in flexibility and performance perspectives over RL, BO, and gradient-based methods \cite{Ref:01,Ref:26,Ref:31,Ref:10}, it is the basis of our PRE-NAS. But unlike other works such as AmoebaNet \citep{Ref:08} which uses aging evolution, PRE-NAS adopts an elitism strategy. Note that, EA algorithms in NAS usually manipulate and update only a few networks in a population as the cost of operating on all networks can be prohibitive \cite{Ref:54}.

The details are shown in \textbf{Algorithm \ref{algo}}. Steps 4-8 comprise the initialisation phase which prepares the population by randomly generating $P$ architectures (Step 5) and using conventional training to obtain validation accuracy on the target task (Step 6).  A $model$ contains architecture and its accuracy will be added into the $population$ and $history$ (Steps 7\&8).   The training samples for the performance predictor are stored in the $history$.  After the initialisation, the search begins. The key differences compared to other EAs are: (1) multi-mutation is used to generate a group of child networks (Step 14); (2) the performance predictor is used to predict the accuracy of child networks (Step 16) which allows us to evaluate many architectures with a much lower cost; (3) a representative selection strategy is adopted to pick distinctive architectures (Step 19); (4) weight inheritance is used in offspring training (Step 21). The Spearman coefficient is calculated between predicted and trained accuracy (Step 24).  A high Spearman coefficient indicates the predictor can better predict the tendency of the child networks.
 
  \begin{figure}[!h]
    \begin{center}
  	\includegraphics[width=\linewidth,height=0.17\textheight]{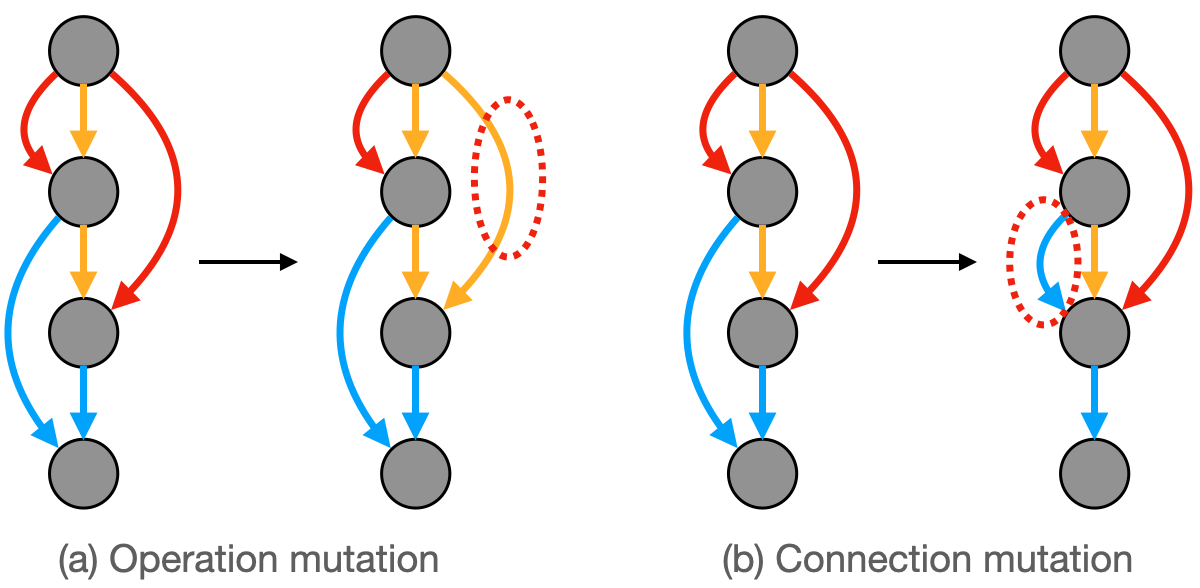}
  	\caption{Illustration of the two mutation strategies of PRE-NAS.  Grey circles represent nodes of the cell network and coloured lines represent different network operations. (a) An offspring network generated by operation mutation, the red dashed circle indicates that the network operation at this position has been mutated. (b) An offspring network generated by connection mutation, the red dash line circle indicates the connection between two nodes has been mutated.}
  	\label{mutation}
  	\end{center}
\end{figure} 

\subsection{Multi-mutation}
\label{sec-mulmutation}
Fig. \ref{mutation} shows two mutation operators in PRE-NAS. Each graph is a cell network and a grey circle is a network node.  Coloured lines represent different network operations, e.g. convolution or pooling.  An arrow represents the direction of the tensor flow. The operation mutation (Fig. \ref{mutation}a) generates an offspring network from a parent network by randomly choosing a network operation and replacing it with another randomly picked operation.  The connection mutation (Fig. \ref{mutation}b) is similar but changes the destination node of an operation.  Theoretically, it is possible to explore the entire space by alternating these two mutation operators. 
The AmoebaNet \cite{Ref:08} suffers from high computational overhead, as it only generates one offspring by mutating the parent in each search cycle. In PRE-NAS, a multi-mutation strategy is used to take advantage of the predictor so all possible offspring networks of a given parent can be generated and evaluated in each search cycle to better cover the search space. The total number of possible offsprings from a parent network can be calculated by $2N(O-1)+2N!$, where N represents the number of intermediate nodes and O represents the number of network operations. As discussed in Section \ref{search_space}, the DARTS space consists of 4 intermediate nodes and 8 operations, which means a parent network can generate up to 76 different offsprings.


\subsection{Representative Selection}
After mutation, the performances of offspring networks will be predicted by our performance predictor. Good candidate networks can be picked by the top-$k$ strategy \cite{Ref:40,Ref:31,Ref:41}.  However, this strategy is not adequate to cover the entire distribution of different scenarios, thus, it is hard to train a predictor with good generalisation. Besides, it needs a way to determine the optimal $K$.  To better maintain diversity and coverage, we propose a representative selection strategy that selects representative architectures from the $children$, not just the best ones.  More specifically, the architectures are selected based on the statistical percentile of predicted validation accuracy: the architectures at the maximum percentile, 75\% point, 50\% point, 25\% point, and the minimum percentile. After the selection, the ground-truth accuracy of these representative candidates can be obtained by training them on the target dataset. The Spearman Ranking Correlation Coefficients (a monotonic relationship between two groups) are calculated between their predicted accuracy and ground-truth accuracy of the representative architectures. The higher the Spearman value, the better ranking of the predictor. Further, the representative architectures and their accuracy values will be added to the $history$ to be the samples for further training of the performance predictor. More exhibitions and discussions on the advantages of our representative selection strategy will be introduced in Section \ref{search_on_nasbench201}.

\subsection{Performance Predictor}
\label{sec-predictor}
The desired characteristics of a PRE-NAS predictor are: (1) it can learn from limited training samples, as the cost of obtaining the target value (validation accuracy) is expensive; (2) it contains only a few model parameters, as training a big model will slow down the overall search process.
The encoding scheme described in the Section \ref{search_space}, architectures are encoded as upper triangular adjacency matrices. Multi-mutations are applied to this encoding, which will also be the input to the performance predictor.  Note that architectures from different search spaces would have matrices of different sizes.  For instance, architectures from the search space of NAS-Bench-201 will be encoded as a $4 \times 4$ matrix, architectures from DARTS will be encoded as an $11 \times 11$ matrix (Fig. \ref{encoding}).

The performance predictors we considered are Random Forest Regressor, Support Vector Regressor, Bayesian Ridge Regressor, Kernel Ridge, Linear Regressor, and Multi-Layer Perceptron, due to their computational efficiency.  These predictors are tested on the NAS-Bench-201.  
To simulate the evaluation process during the architecture search, we randomly sample 100 architectures and their accuracy as the training data (less than 1\% of the total amounts of the search space) and randomly sample another 100 pairs from the rest of 15525 architectures as the test data.  We record the Spearman coefficient between predicted and ground-truth accuracy.
500 independent experiments show that Random Forrest is the best performing and most consistent predictor when facing extremely limited training samples.  The results are shown in Table \ref{predictors}.

\begin{table}[!h]
\centering
\begin{tabular}{l|c}
\hline
\textbf{Predictor}     & \textbf{Spearman Coefficient} \\ \hline
Random Forest          & 0.65±0.08                     \\ \hline
Support Vector         & 0.39±0.11                     \\ \hline
Bayesian Ridge         & 0.04±0.11                     \\ \hline
Kernel Ridge           & 0.06±0.10                     \\ \hline
Linear Regressor       & 0.04±0.11                     \\ \hline
Multi-Layer Perceptron & 0.06±0.10                     \\ \hline
\end{tabular}
\caption{Results of 6 regressors trained with 100 samples.  Each regressor is trained with the same data in each experiment.  We record the Spearman coefficient between predicted and ground-truth accuracy, and report the mean and standard deviation based on 500 independent experiments.}
\label{predictors}
\end{table}

\subsection{High-Fidelity Weight Inheritance}
\label{sec:hifi-WI}

\begin{figure}[!h]
    \begin{center}
  	\includegraphics[width=\linewidth,height=0.16\textheight]{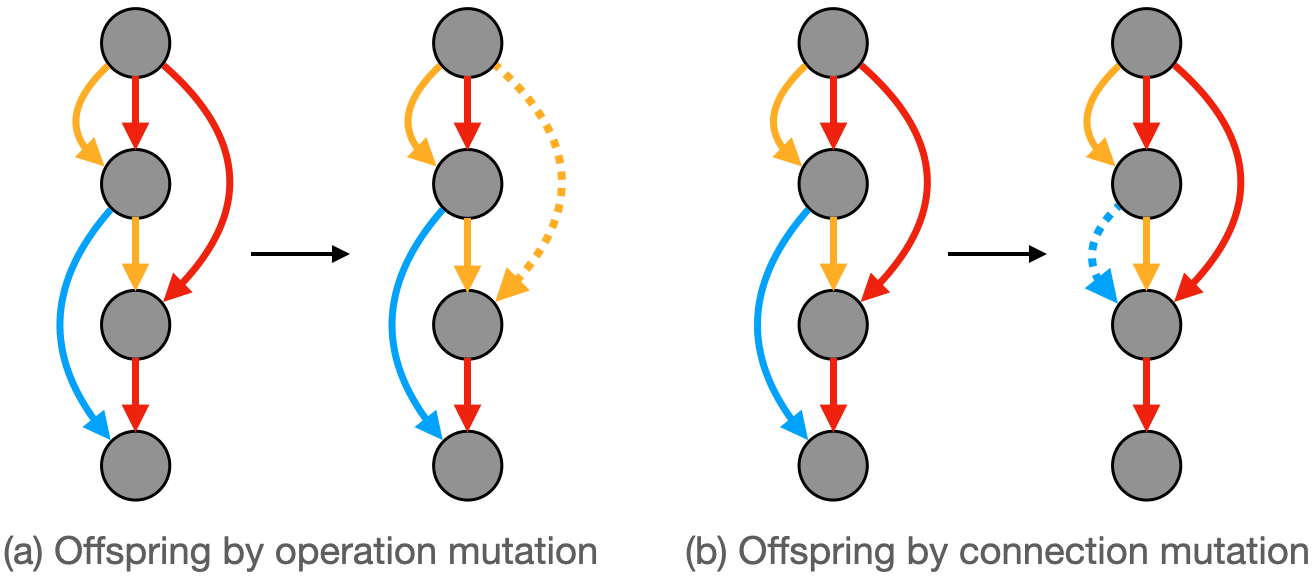}
  	\caption{Illustration of weight inheritance, grey circles represent nodes of the cell network, and colour lines represent different network operations. (a) and (b) demonstrate two offspring networks generated by operation and connection mutations. Solid lines indicate that these operations will inherit the weights from the parent network.  Dash lines indicate that the weights of these operations will reset.}
  	\label{weightinherit}
  	\end{center}
\end{figure}

Similar to the well known one-shot model \cite{Ref:10,Ref:36}, weight inheritance is also based on the weight sharing technique.  One-shot allows sharing the weights across the entire search space, even for heterogeneous architectures.  Weight inheritance however only shares weights between topologically homogeneous networks, e.g., parent and offspring networks.  Thus, evolutionary search is naturally suitable for the weight inheritance training, because offspring networks are highly similar in terms of topology as they are generated from parent networks by mutating particular operations or connections (Fig. \ref{weightinherit}).  Unlike previous work on the variable size of networks \cite{Ref:55,Ref:36}, we adopt a fixed size weight inheritance strategy to best preserve the network function.  Moreover, we use Kaiming Norm \cite{Ref:56} to initialise the mutated connection or operation weights.

To investigate the effectiveness of our weight inheritance, we generated 350 candidate networks by implementing multi-mutation on randomly generated parent networks from the search space of DARTS. Both conventional (from scratch) and weight inheritance training were employed on the CIFAR-10 dataset.  Both training runs were optimised with momentum SGD, and a batch size of 128. 
The hyper-parameters for the former strategy were set as follows: learning rate was 0.025 (annealed via cosine strategy), momentum was 0.9, weight decay was 0.0001, and number of training epochs was 100. 
For weight inheritance training, as offsprings will partially inherit trained weights from parent networks, both learning rate and training epochs were reduced as follow: the learning rate was set to 0.01 and the number of training epochs was set to 50.

 \begin{figure}[t!]
    \begin{center}
  	\includegraphics[width=0.9\linewidth,height=0.32\textheight]{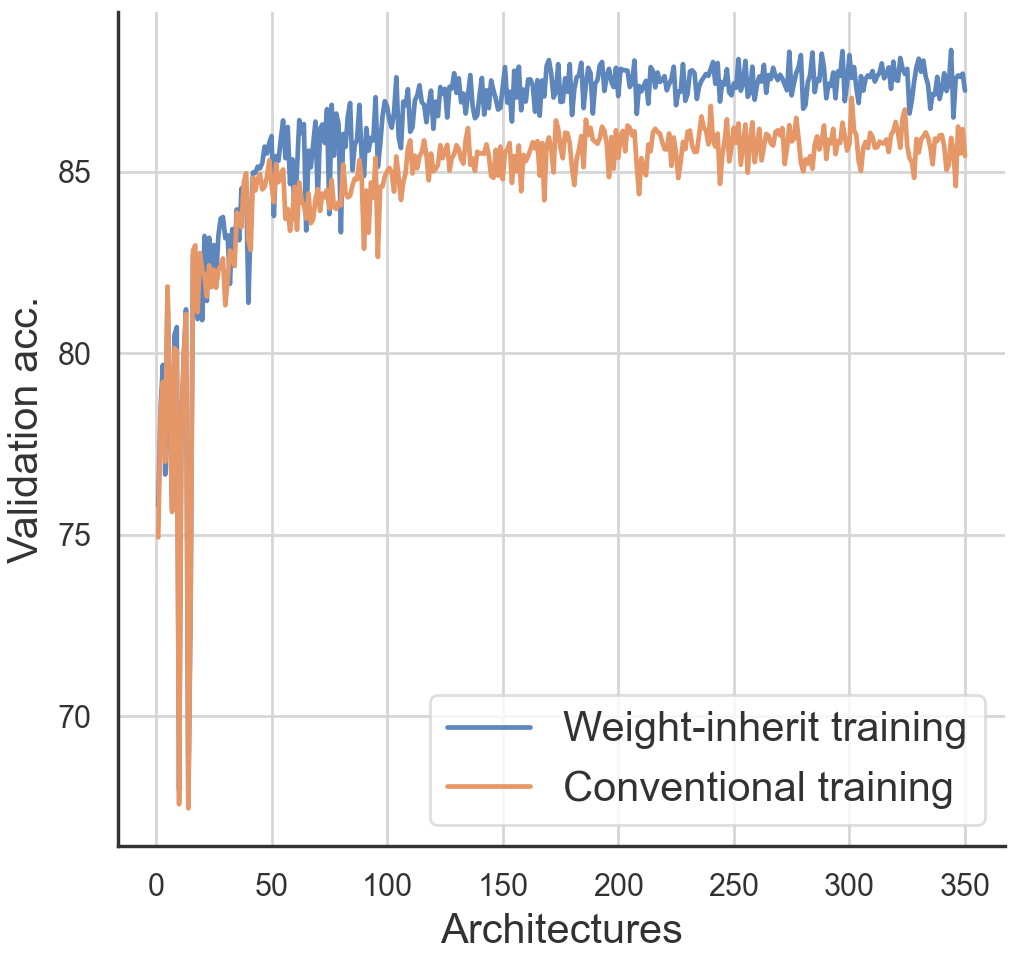}
  	\caption{Results of weight inheritance and conventional training on 350 candidate networks.  The network order is sorted by the CIFAR-10 validation accuracy.  The Spearman Correlation Coefficient between these two groups is 0.83.}
  	\label{c_w_compare}
  	\end{center}
\end{figure} 

Fig. \ref{c_w_compare} shows the CIFAR-10 validation accuracy of these 350 networks obtained by (1) weight inheritance and (2) conventional training.  The Spearman Coefficient between these two groups is 0.83, which indicates the high fidelity of inherited weights. This is significantly better than the one-shot model that shows poor correlation with the ground-truth accuracy \cite{Ref:50}.  The weight inheritance strategy can obtain both training efficiency and high ranking correlation, as it required much fewer training epochs to reach a similar performance when compared with conventional training.

\section{Experiments and Results}
\label{section4}
We have implemented our experiments on two search spaces which are NAS-Bench-201 \cite{Ref:45} and DARTS \cite{Ref:10}. We use NAS-Bench-201 space to implement a large scale of experiments, it aims to prove the stability of search performance of PRE-NAS. We use DARTS space to implement real-world architecture search, which can generate high-performance neural networks.

\begin{table*}[!h]
\centering
\begin{tabular}{l|c|c|c|c|c|c}
\hline
\textbf{Algorithm} & \multicolumn{2}{c|}{\textbf{CIFAR-10}} & \multicolumn{2}{c|}{\textbf{CIFAR-100}} & \multicolumn{2}{c}{\textbf{ImageNet-16-120}} \\ \hline
                                        & Validation                     & Test                          & Validation                     & Test                           & Validation                        & Test                              \\ \hline
RSPS \cite{Ref:32}                                    & 84.16±1.69                     & 87.16±1.69                    & 59.00±4.60                     & 58.33±4.34                     & 31.56±3.28                        & 31.14±3.88                        \\
DARTS-V1 \cite{Ref:10}                               & 39.77±0.00                      & 54.30±0.00                     & 15.03±0.00                      & 15.61±0.00                      & 16.43±0.00                         & 16.32±0.00                         \\
DARTS-V2 \cite{Ref:10}                       & 39.77±0.00                      & 54.30±0.00                     & 15.03±0.00                     & 15.61±0.00                      & 16.43±0.00                         & 16.32±0.00                         \\
GDAS \cite{Ref:63}                        & 90.00±0.21                     & 93.51±0.13                    & 71.14±0.27                     & 70.61±0.26                     & 41.70±1.26                        & 41.84±0.90                        \\
SETN \cite{Ref:64}                             & 82.25±5.17                     & 86.19±4.63                    & 56.86±7.59                     & 56.87±7.77                     & 32.54±3.63                        & 31.90±4.07                        \\
ENAS \cite{Ref:17}                                    & 39.77±0.00                      & 54.30±0.00                     & 15.03±0.00                      & 15.61±0.00                      & 16.43±0.00                         & 16.32±0.00                         \\ 
AmoebaNet \cite{Ref:08}                                     & 91.19±0.31                     & 93.92±0.30                    & 71.81±1.12                     & 71.84±0.99                     & 45.15±0.89                        & 45.54±1.03                        \\
Random Search \cite{Ref:65}                                      & 90.93±0.36                     & 93.70±0.36                    & 70.93±1.09                     & 71.04±1.07                     & 44.45±1.10                        & 44.57±1.25                        \\
REINFORCE \cite{Ref:01}                              & 91.09±0.37                     & 93.85±0.37                    & 71.61±1.12                     & 71.71±1.09                     & 45.05±1.02                        & 45.24±1.18                        \\
BOHB \cite{Ref:66}                                    & 90.82±0.53                     & 93.61±0.52                    & 70.74±1.29                     & 70.85±1.25                     & 44.26±1.36                        & 44.42±1.49                        \\ \hline
PRE-NAS (ours)                                 & {91.37±0.28}                     & {94.04±0.34}                    & {71.95±1.21}                     & {72.02±1.22}                     & {45.16±1.00}                        & {45.34±1.03}                        \\\hline
\textbf{*Optimal}                        & 91.61                          & 94.37                         & 73.49                          & 73.51                          & 46.77                             & 47.31                             \\ \hline
\end{tabular}
\caption{Search results of our proposed PRE-NAS and other search algorithms on the NAS-Bench-201 search space. Each algorithm is searching under a similar computational budget and repeated 500 times. We recorded the CIFAR-10 validation accuracy of the best-performing architecture which found in each experiment and reported the mean and standard deviation over 500 independent experiments. \textbf{*Optimal} indicates the accuracy of best-performing architecture recorded in the benchmark.}
\label{nasbench201results}
\end{table*}

\subsection{Search on NAS-Bench-201}
\label{search_on_nasbench201}
As the search space of NAS-Bench-201 has been exhaustively evaluated and the performance of every network within the search space is known, there is no need for training the initial population or offspring candidates, but query the record.  On this benchmark, PRE-NAS was repeated 500 times in order to reduce the variance between runs. The computational budget was set according to the experiments in \cite{Ref:45}. We set the search $Cycle \; C$ to 20,  $Population\_size \; P$ to 20, $Sample\_size \; S$ to 10, and initial $mutation\_times$ equal to the $Sample\_size \; S$.  This means that the algorithm runs mutation operations 10 times to generate 10 offspring networks.  Once the child networks are evaluated by the performance predictor, then the percentile selection will apply.  There is one additional hyper-parameter called $mutation\_factor$, which is used to increase $mutation\_times$.  For instance, if the predictor performed well in the current search cycle, $mutation\_times$ will be increased by $mutation\_factor$.  Thus, the predictor could gradually evaluate more architectures if it performs well.  This parameter was set to 1.2.  This setup ensures a similar computational budget which is the total number of architectures that have been queried by using the benchmark. Note that the total number of training samples for the performance predictor is around 100 under this setup (approx. $10^{-4}$ of the search space), which is similar to our simulation experiment shown in Section \ref{sec-predictor}.

\subsubsection{NAS-Bench-201 Results}
Table \ref{nasbench201results} shows the results from PRE-NAS and respective SOTA algorithms over 500 independent experiments.  The columns under CIFAR-10, CIFAR-100 and ImageNet-16-120 indicate the average validation and test accuracy with the standard deviation of the best architecture by each algorithm.  The last row shows the best accuracy that the architecture in this search space could possibly reach. The large scale of independent architecture search experiments clearly show that the overall performance of PRE-NAS dominates all other algorithms, given a similar computational budget.

 \begin{figure}[!h]
    \centering
  	\includegraphics[width=0.77\linewidth,height=0.24\textheight]{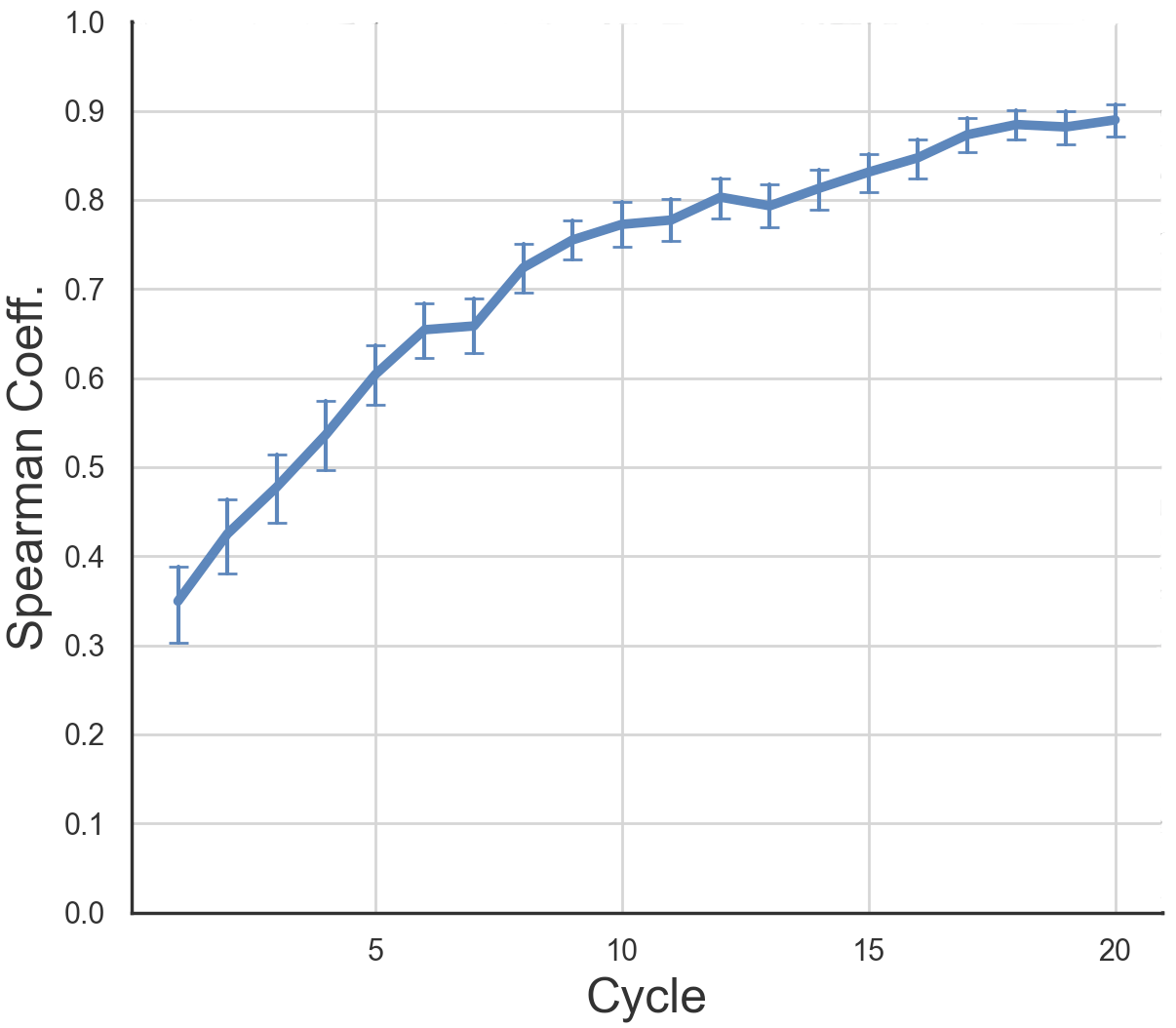}
  	\caption{Result of Spearman coefficient between predicted and ground-truth accuracy. The line indicates the average value of the Spearman coefficient at each search cycle based on 500 independent experiments.
  	}
  	\label{spearman201}
\end{figure}

\begin{figure}[!h]
\begin{subfigure}{.45\textwidth}
  \centering
  \includegraphics[width=0.8\linewidth,height=0.25\textheight]{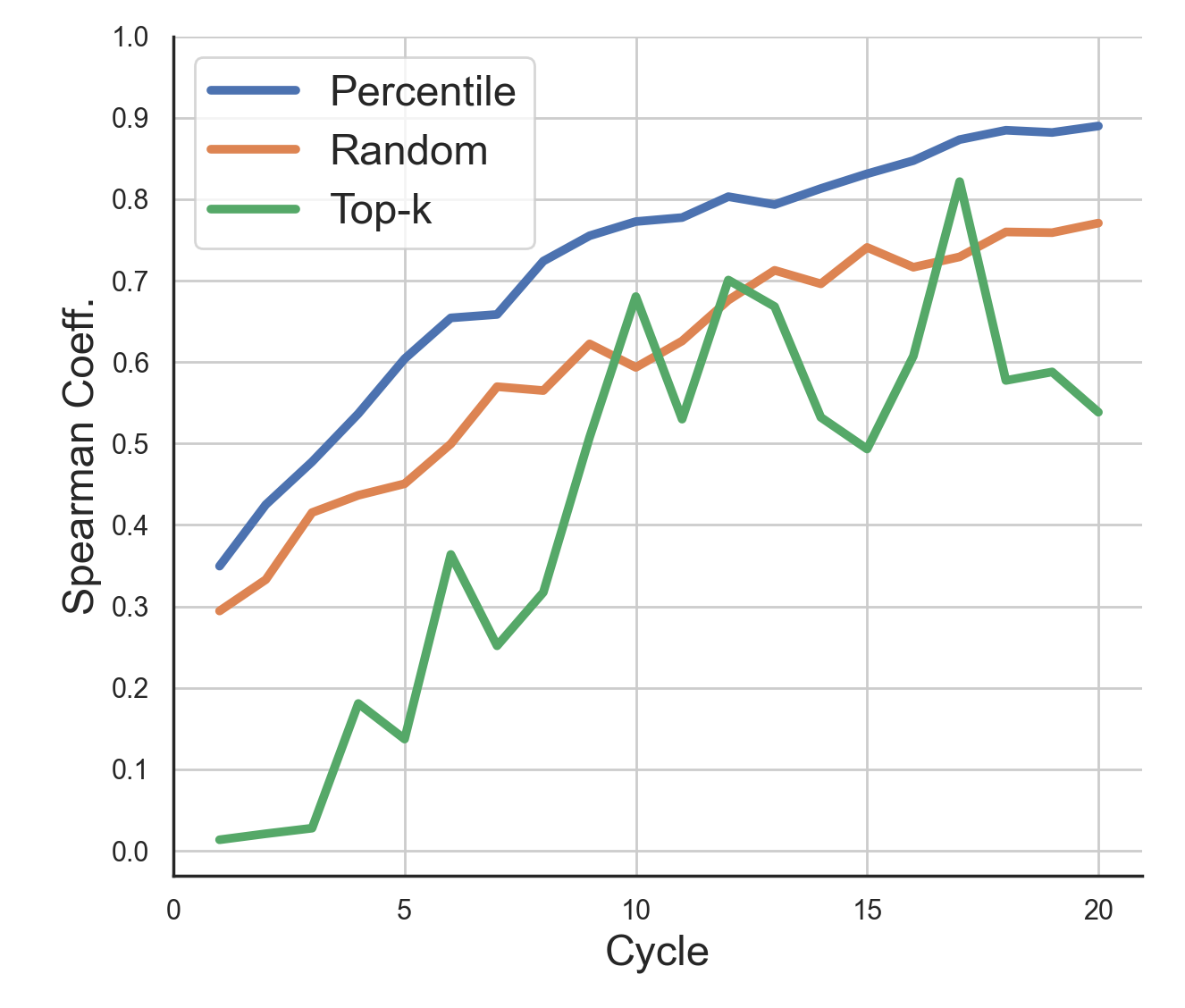}
  \caption{}
  \label{strategy_spr}
  \end{subfigure}
  \begin{subfigure}{.45\textwidth}
  \centering
  \includegraphics[width=0.83\linewidth,height=0.25\textheight]{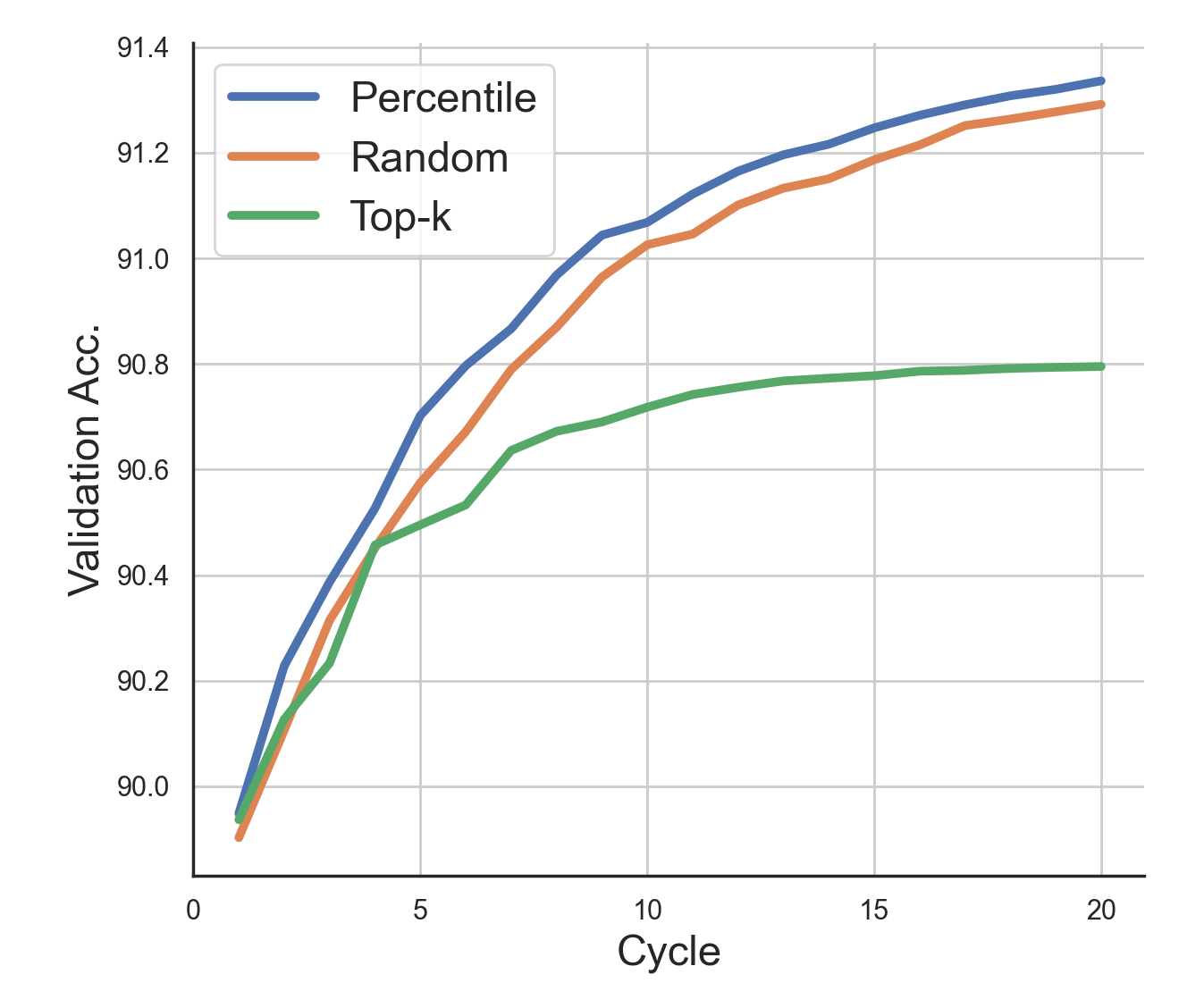} 
  \caption{}
  \label{strategy_acc}
  \end{subfigure}
\caption{(a) Spearman coefficient between predicted and ground-truth accuracy under three candidates selection strategies. (b) Ground-truth accuracy under three candidates selection strategies. Each line indicates the average value at each search cycle based on 500 independent experiments.
}
\label{strategy_compare}
\end{figure}



We also recorded the Spearman coefficient between predicted and ground-truth CIFAR-10 validation accuracy at each search cycle.  Fig. \ref{spearman201} shows the average value of the Spearman coefficient over 500 independent experiments.  Even with extremely limited training samples, the performance predictor was still continuously improving, taking advantage of multi-mutation and representatives selection.  Moreover, we have investigated two extra candidate selection strategies, top-$k$ (e.g. top-5) and random. The experimental setups are the same as for our percentile strategy.  Fig. \ref{strategy_compare} demonstrate that our percentile-based representative selection strategy is superior on both predictor training and architecture search. Based on these results, we think that percentile or even random-based representative selection strategies are more likely to cover the distribution of the search space than the top-$k$ strategy. Thus, it constructed richer training samples for the performance predictor.

\subsection{Search on DARTS}
DARTS search space contains approximately $10^9$ possible architectures, much larger than NAS-Bench-201.  We follow a similar procedure to the experiments on NAS-Bench-201 that use the CIFAR-10 as the proxy task dataset, except that it requires training the networks (shown in \textbf{Algorithm \ref{algo}}).  Specifically, we randomly generate architectures to form the initial population and use conventional training to obtain their accuracy.  Then the offspring networks can inherit weights from them. Therefore we can use weight inheritance training to train the following offspring networks with a much less computational cost. We set the search $Cycle \; C$ to 100, $Population\_size \; P$ to 64, $Sample\_size \; S$ to 32, initial $mutation\_times$ equals the $Sample\_size \; S$, and the $mutation\_factor$ to 1.2. For the initial population, we use conventional training and set the epoch number to 100, momentum to 0.9, learning rate to 0.025 (anneal cosine strategy) and weight decay to 0.0001. We use weight inheritance training for the offspring candidate networks and set the learning rate to 0.01, epoch number to 50 which is the same as the setup in Section \ref{sec:hifi-WI}. Both conventional and weight inheritance training are optimised by momentum SGD and batch size to 128.  We set the network layer to 1 (single cell) and the initial channel to 16 during the architecture search, as training the shallow networks would significantly reduce the computational cost.  Under this setup, there are 64 networks (initial population) trained by the conventional way and approximately 500 networks (offspring networks) trained by weight inheritance. Thus, the theoretical number of training samples for the performance predictor is 564 (approx. $10^{-8}$ of the search space).  PRE-NAS would evaluate around 7000 architectures by implementing the multi-mutation.  Each network would take approximately 10 minutes on a single Tesla V100 GPU if training from scratch. This leads to approximate 48 days to complete the search. By utilising the performance predictor and weight inheritance training, PRE-NAS reduces the computational cost to 0.6 GPU days.

\subsubsection{Result on CIFAR-10}
After PRE-NAS found the promising architecture, we use it to construct a full-size network as discussed in Section \ref{search_space}. For the training of our full-size network, we apply a similar setup with DARTS \cite{Ref:10}, i.e., stacked the cell network with 20 layers. Table \ref{cifar10results} shows the results of the comparison between our full-size network and the networks from other methods. As the CIFAR-10 results have high variance between different runs, we trained our full-size model 10 times and reported the mean and standard deviation. From the results, the architecture found by PRE-NAS is significantly better than most state-of-the-art which also searched on DARTS space. PRE-NAS only took 0.6 GPU days for the search. From the perspective of search cost, our method is still not inferior to the mainstream one-shot NAS. In evolution-based search algorithms, PRE-NAS also shows a huge advantage in both search efficiency and performance.

\begin{table*}[!th]
\centering
\begin{tabular}{l|c|c|c|c|c|c|c}
\hline
\textbf{Algorithm} & \textbf{\begin{tabular}[c]{@{}c@{}}Test Error\\ (\%)\end{tabular}} & \textbf{\begin{tabular}[c]{@{}c@{}}Params\\ (M)\end{tabular}} & \multicolumn{1}{l}{\textbf{\ \ \ \ \ GPU}} & \textbf{\begin{tabular}[c]{@{}c@{}}Search Cost\\ (GPU Days)\end{tabular}} & \multicolumn{1}{l}{\textbf{Search Method}} & \begin{tabular}[c]{@{}c@{}}{\textbf{Evaluation}}\end{tabular} & \begin{tabular}[c]{@{}c@{}}{\textbf{Year}}\end{tabular} \\ \hline
ENAS \cite{Ref:17}                                       & 2.89$\star$                                                                                         & 4.6                                                           & -                                & 0.45                                                                      & Reinforce                                  & One-shot   &ICML2018                                    \\ 
PNAS \cite{Ref:40}                                      & 3.34±0.09                                                                                  & 3.2                                                           & -                                & 225                                                                       & SMBO                                       & Predictor     &ECCV2018                                 \\
RandomNAS \cite{Ref:32}$\dagger$                                      & 2.85±0.08                                                                                      & 4.3                                                           & P100\&V100                               & 2.7                                                                       & Random                                  & One-shot &UAI2020                                       \\
DARTS \cite{Ref:10}$\dagger$                         & 3.00±0.14                                                                                & 3.3                                                           & 4 1080Ti                         & 4                                                                       & Gradient                                   & One-shot        &ICLR2019                               \\
P-DARTS \cite{Ref:52}$\dagger$                                       & 2.50$\star$                                                                                        & 3.4                                                           & -                                & 0.3                                                                       & Gradient                                  & One-shot  &ICCV2019                                     \\
FairDARTS \cite{Ref:35}$\dagger$                                       & 2.54$\star$                                                                                        & 2.8                                                           & 1 V100                                & 0.42                                                                       & Gradient                                  & One-shot   &ECCV2020                                    \\
CGP-ResSet \cite{Ref:59}                               & 5.01$\star$                                                                                    & 1.7                                                           & 2 1080Ti                          & 4                                                                      & Evolution                                  & Conventional         &IJCAI2018                          \\
AmoebaNet-A \cite{Ref:08}                               & 3.34±0.06                                                                                  & 3.2                                                           & 450 K40                          & 3150                                                                      & Evolution                                  & Conventional         &AAAI2019                          \\
Lemonade \cite{Ref:60}                               & 3.05$\star$                                                                                   & 4.7                                                           & 16 Titan X                          & 56                                                                      & Evolution                                  & Conventional         &ICLR2019                          \\
EENA \cite{Ref:51}                               & 2.56$\star$                                                                                   & 8.47                                                           & 1 Titan X                          & 0.65                                                                      & Evolution                                  & Weight Inheritance         &ICCV2019                          \\
NSGA-Net \cite{Ref:61}                               & 2.75$\star$                                                                                   & 3.3                                                           & 1 1080Ti                          & 4                                                                      & Evolution                                  & Predictor         &GECCO2019                          \\
NSGANetV1-A4 \cite{Ref:61}                               & 2.02$\star$                                                                                   & 4.0                                                           & 8 2080Ti                          & 27                                                                      & Evolution                                  & Predictor         &TEC2020                          \\
EcoNAS \cite{Ref:58}$\dagger$                                       & 2.62±0.02                                                                                      & 2.9                                                           & 1 1080Ti                                & 8                                                                       & Evolution                                  & Conventional  &CVPR2020                                     \\

CARS \cite{Ref:07}$\dagger$                                       & 2.62$\star$                                                                                       & 3.6                                                           & -                                & 0.4                                                                       & Evolution                                  & One-shot       &CVPR2020 \\
EvNAS \cite{Ref:62}$\dagger$                                       & 2.47±0.06                                                                                     & 3.6                                                           & 1 2080 Ti                                & 3.83                                                                       & Evolution                                  & One-shot       &GECCO2021 \\\hline

PRE-NAS (ours)$\dagger$                                       & 2.49±0.09                                                                                      & 4.5                                                           & 1 V100                             & 0.6                                                                         & Evolution                                  & Predictor+w/i                                      \\ \hline
\end{tabular}
\caption{Performance comparison between the networks found by PRE-NAS and other search algorithms on CIFAR-10, the lower test error rate is better. $\dagger$ means the method has also been used on DARTS search space.  $\star$ indicates the original paper only provided their best performance. Dash means the original paper has not provided the information. `w/i' is short for the weight inheritance training.}
\label{cifar10results}
\end{table*}

\subsubsection{Result on ImageNet}
As the ImageNet dataset contains approximate 13k large size images with 1000 classes, directly searching on ImageNet will cause gigantic computational overhead. The performance transferability of cell networks have been proved in most of the previous work \cite{Ref:10,Ref:21,Ref:35,Ref:50}. Our full-size network for ImageNet is also constructed from the architecture found on CIFAR-10, we apply a similar setup with DARTS \cite{Ref:10} to build a slightly larger model. Table \ref{imagenetresults} shows the results of comparison between our model and the networks from other work, which is consistent with the results shown in Table \ref{cifar10results}.

\begin{table*}[!th]
\centering
\begin{tabular}{l|c|c|c|c|c|c|c}
\hline
\textbf{Algorithm} & \textbf{\begin{tabular}[c]{@{}c@{}}Test Error(\%) \\ Top-1 / Top-5 \end{tabular}} & \textbf{\begin{tabular}[c]{@{}c@{}}Params\\ (M)\end{tabular}} & \multicolumn{1}{l} {\textbf{\ \ \ GPU}} & \textbf{\begin{tabular}[c]{@{}c@{}}Search Cost\\ (GPU Days)\end{tabular}} & \multicolumn{1}{l}{\textbf{Search Method}} & \begin{tabular}[c]{@{}c@{}}{\textbf{Evaluation}}\end{tabular} & \begin{tabular}[c]{@{}c@{}}{\textbf{Year}}\end{tabular}\\ \hline 
                                      
PNAS \cite{Ref:40}                                       & 25.8 / 8.1                                                                                  & 5.1                                                           & -                                & 225                                                                       & SMBO                                       & Predictor   &ECCV2018                                    \\
DARTS \cite{Ref:10}$\dagger$                         & 26.7 / 8.7                                                                                & 4.7                                                           & 4 1080Ti                         & 4                                                                       & Gradient                                   & One-shot  &ICLR2019                                     \\
P-DARTS \cite{Ref:52}$\dagger$                     & 24.4 / 7.4                                                                                         & 4.9                                                           & -                                & 0.3                                                                       & Gradient                                  & One-shot &ICCV2019                                        \\
FairDARTS \cite{Ref:35}$\dagger$                  & 26.3 / 8.3                                                                                        & 5.3                                                           & 1 V100                                & 0.42                                                                       & Gradient                                  & One-shot           &ECCV2020                            \\
AmoebaNet-A \cite{Ref:08}                        & 25.5 / 8                                                                                  & 5.1                                                           & 450 K40                          & 3150                                                                      & Evolution                                  & Conventional  &AAAI2019                                  \\
EcoNAS \cite{Ref:58}$\dagger$                  & 25.2 / -                                                                                      & 4.3                                                           & 1 1080Ti                                & 8                                                                       & Evolution                                  & Conventional  &CVPR2020                                        \\

CARS \cite{Ref:07}$\dagger$                  & 24.8 / 7.5                                                                                       & 5.1                                                           & -                                & 0.4                                                                       & Evolution                                  & One-shot  &CVPR2020                                     \\
NSGANetV1-A3 \cite{Ref:61}                   & 23.8 / 7                                                                                   & 5.0                                                           & 8 2080Ti                          & 27                                                                      & Evolution                                  & Predictor         &TEC2020                          \\
EvNAS \cite{Ref:62}$\dagger$              & 24.4 / 7.4                                                                                     & 5.1                                                           & 1 2080 Ti                                & 3.83                                                                       & Evolution                                  & One-shot       &GECCO2021 \\\hline

PRE-NAS (ours)$\dagger$                    & 24 / 7.8                                                                                       & 6.2                                                           & 1 V100                             & 0.6                                                                         & Evolution                                  & Predictor+w/i                                      \\ \hline
\end{tabular}
\caption{Performance comparison between the networks found by PRE-NAS and other search algorithms on the ImageNet (mobile setting), the lower test error rate is better.  
$\dagger$ means the method has also been used on DARTS search space. Dash means the original paper has not provided the information. `w/i' is short for the weight inheritance training.}
\label{imagenetresults}
\end{table*}

\section{Conclusion}
In this paper, we proposed a predictor-assisted evolutionary search algorithm called PRE-NAS. We demonstrated that the predictor-based evolutionary NAS is highly competitive with mainstream methods. By adopting the multi-mutation and representative selection strategies, we can train a good performance predictor with extremely limited data (e.g. approx. $10^{-8}$ to $10^{-4}$ of the total size of the search space).  The predicted accuracy shows a strong correlation with the ground-truth accuracy. By utilising the high-fidelity weight inheritance strategy, we can further reduce the cost of candidate network training.  On DARTS search space, the search cost is reduced from 48 GPU days to 0.6 GPU days. Extensive experiments show that the network found by PRE-NAS can outperform several state-of-the-art search algorithms on the search spaces of NAS-Bench-201 and DARTS. 

In this work, we only used network accuracy as the search objective.  Introducing multi-objectives could be a promising direction for future work as multiple requirements, e.g. accuracy and model size, then can be considered simultaneously during architecture search.  Furthermore, it can be observed that the performance of evolutionary algorithms depends on the quality of the initial population.  A high-quality initial population is more likely to generate good offspring networks.  In near future, we will investigate ways to better initialize the initial populations. 

\bibliographystyle{ACM-Reference-Format}
\bibliography{sample-base}


\end{document}